\documentclass[sigconf]{acmart}

\AtBeginDocument{%
  }

\usepackage{hyperref}
\usepackage{multirow}
\usepackage{todonotes}
\usepackage{tabularx}
\setcopyright{acmlicensed}
\copyrightyear{2018}
\acmYear{2018}
\acmDOI{XXXXXXX.XXXXXXX}

\acmConference[JCDL '24]{ACM/IEEE-CS Joint Conference on Digital Libraries}{December 16--20,
  2024}{Hong Kong, China}
\acmISBN{978-1-4503-XXXX-X/18/06}

\begin{document}



\title{LLMs4Synthesis: Leveraging Large Language Models for Scientific Synthesis}



\author{Hamed Babaei Giglou}
\affiliation{
  \institution{TIB  Leibniz Information Centre for\\ Science and Technology}
  \city{Hannover}
  \country{Germany}}
\email{hamed.babaei@tib.eu}

\author{Jennifer D'Souza}
\affiliation{
  \institution{TIB  Leibniz Information Centre for\\ Science and Technology}
  \city{Hannover}
  \country{Germany}}
\email{jennifer.dsouza@tib.eu}
 
\author{S\"{o}ren Auer}
\affiliation{
  \institution{TIB  Leibniz Information Centre for\\ Science and Technology}
  \city{Hannover}
  \country{Germany}}
\email{auer@tib.eu}

\renewcommand{\shortauthors}{Author et al.}

\begin{abstract}
In response to the growing complexity and volume of scientific literature, this paper introduces the LLMs4Synthesis framework, designed to enhance the capabilities of Large Language Models (LLMs) in generating high-quality scientific syntheses. This framework addresses the need for rapid, coherent, and contextually rich integration of scientific insights, leveraging both open-source and proprietary LLMs. It also examines the effectiveness of LLMs in evaluating the integrity and reliability of these syntheses, alleviating inadequacies in current quantitative metrics. Our study contributes to this field by developing a novel methodology for processing scientific papers, defining new synthesis types, and establishing nine detailed quality criteria for evaluating syntheses. The integration of LLMs with reinforcement learning and AI feedback is proposed to optimize synthesis quality, ensuring alignment with established criteria. The LLMs4Synthesis framework and its components are made available, promising to enhance both the generation and evaluation processes in scientific research synthesis.
\end{abstract}

\ccsdesc[300]{Information systems~Information systems applications}
\ccsdesc[500]{Computing methodologies~Natural language generation}
\ccsdesc[500]{Computing methodologies~Language resources}
\ccsdesc[300]{Computing methodologies~Natural language processing}
\ccsdesc[500]{Computing methodologies~Supervised learning}
\ccsdesc[500]{Computing methodologies~Reinforcement learning}

\keywords{Scientific Synthesis Generation, Large Language Models, Language Model-based Evaluation Framework, Reinforcement Learning}

\received{8 August 2024}
\received[revised]{12 March 2009}
\received[accepted]{5 June 2009}

\maketitle

\section{Introduction}

In recent years, the intersection of language processing and scientific research has witnessed major advancements, largely fueled by the capabilities of Large Language Models (LLMs) \cite{llmcatalog,bubeck2023sparks}. These models, including open-source (e.g. BERT~\cite{devlin2019bert} or Mistral~\cite{jiang2023mistral7b}), or proprietary technologies (e.g. GPT-4~\cite{achiam2023gpt4}), and their derivatives, have not only redefined the boundaries of text generation and comprehension~\cite{srivastava2023beyond} but have also paved the way for innovative applications in the synthesis of scientific content~\cite{core-gpt,evans2024large}. To enable the automated and accurate generation of scientific syntheses, which are concise integrations of key insights from multiple scientific articles, this paper introduces the comprehensive LLMs4Synthesis framework. It is designed to enhance open-source LLMs so that they can generate scientific syntheses of comparable quality to those produced by significantly larger proprietary models.

The motivation behind this endeavor is driven by the increasing complexity and volume of scientific literature~\cite{VanNoorden2014,bornmann2015growth,fortunato2018science}, which poses challenges for researchers seeking timely and comprehensive knowledge syntheses. Traditional manual synthesis methods are time-consuming and struggle to keep up with the rapid dissemination of new research. The LLMs4Synthesis framework addresses these challenges by leveraging LLMs to produce concise, coherent, and contextually rich scientific summaries, helping the scientific community stay abreast of expanding knowledge frontiers. Furthermore, next-generation search engines like \href{https://ask.orkg.org/}{ORKG Ask}~\cite{orkgask2024}, \href{https://elicit.com/}{Elicit}~\cite{elicit2024}, and \href{https://typeset.io/}{SciSpace}~\cite{typeset2024} demonstrate how LLMs enhance search capabilities and user interactions. These platforms utilize advanced features such as natural language queries, semantic search, and AI-driven extractions to transform scientific literature access. ORKG Ask is an open platform, allowing for seamless integration of LLMs4Synthesis, enhancing synthesis abilities, and promoting open access and collaboration in the scientific community.

Accurate evaluation of scientific syntheses is essential to maintain their integrity and reliability. Recent advancements suggest that LLMs can generate these syntheses \cite{core-gpt,orkgask2024,elicit2024,typeset2024}, yet their effectiveness across domains and capability in evaluating them is not well understood. 
Merely employing existing quantitative metrics, like the \textsc{rouge} family \cite{lin-2004-rouge}, commonly applied in similar text generation scenarios, raise concerns about their adequacy \cite{cohan-goharian-2016-revisiting,kryscinski-etal-2019-neural,akter-etal-2022-revisiting,evans2024large} in capturing the full extent of the semantic intricacies associated with the quality evaluation of syntheses. Instead, our research examines the effectiveness of LLMs in assessing scientific syntheses, contributing to a significant methodological shift where language understanding tools are increasingly observed as effective evaluators \cite{agastya2023decoding,llm-examiner}. 
Moreover, related work \cite{llm-as-a-judge,llm-eval,fu2023gptscore} shows that using LLMs as evaluators facilitates intricate versatility and detail in assessments across multiple dimensions and scales.
Consequently, our study aims to define a comprehensive set of fine-grained evaluation attributes to address critical synthesis issues like irrelevancy, inaccuracy, incompleteness, redundancy, disorganization, incoherence, poor readability, and verbosity, which significantly affect the quality and clarity of syntheses.
Additionally, this study explores using LLM evaluator responses as AI feedback in reinforcement learning settings (RLAIF) \cite{christiano2017deep,bai2022constitutional} to enhance the alignment of a model's unsupervised learned parameters \cite{radford2019language} with the quality of scientific synthesis generation. This approach moves beyond traditional reliance on human evaluators and ground truth data.


In essence, this research aims to equip LLMs with the ability to identify and improve upon their synthesis generation quality, thereby enhancing their learning and output reliability. The study focuses on the following research questions (\textbf{RQs}). \textbf{RQ1:} What features are essential for qualitatively evaluating the informativeness and meaningfulness of a scientific synthesis? \textbf{RQ2:} How can LLMs be fine-tuned to adhere to specific formatting standards like word limits and paragraph structure in scientific syntheses?
\textbf{RQ3:} How can the balance be achieved in LLMs between maintaining formatting standards and ensuring semantically informative and meaningful scientific syntheses?

This paper makes several contributions. First, we develop a methodology to collect and process scientific papers into a format ready for synthesis using the Open Research Knowledge Graph~\cite{auer2020improving}, a multidisciplinary platform that facilitates the comparison of scientific contributions~\cite{OlleenEtAlSciknow19}.
Second, we introduce new synthesis types —- paper-wise, methodological, and thematic —- that focus on different aspects of the extracted insights. Utilizing Mistral-7B \cite{jiang2023mistral7b} and GPT-4 \cite{achiam2023gpt4}, we generate a large-scale dataset of these syntheses, which is publicly available \url{https://github.com/jd-coderepos/scisynthesis}. Third, we establish nine quality criteria for evaluating these syntheses, assessed by both an automated LLM evaluator (GPT-4) and a human-crowdsourced survey. These evaluations provide insights, i.e. at type, domain, LLM levels, in the research community at large, and also help inform the LLMs4Synthesis framework, to enhance LLM scientific synthesis generation task performance. Finally, we propose the LLMs4Synthesis framework, which incorporates RLAIF \cite{christiano2017deep,bai2022constitutional} to optimize LLMs for synthesis generation, ensuring alignment with established quality standards. The framework and its source code are publicly accessible \url{https://github.com/HamedBabaei/LLMs4Synthesis}.

The rest of the paper is organized as follows. Section \ref{sec:synthesis-dataset} details the multidisciplinary ORKG synthesis dataset. Section \ref{sec:quality-eval} introduces comprehensive evaluation attributes and assesses synthesis quality through both automatic and human evaluations, providing performance comparisons between proprietary and open-source LLMs. Section \ref{sec:llms4synth} elaborates on the LLMs4Synthesis framework, with its evaluations in \autoref{sec:llms4syntheval}. Section \ref{sec:discuss} explores the broader implications of our findings, and \autoref{sec:conclude} concludes the paper, summarizing our contributions and suggesting future work.

\section{The ORKG Scientific Synthesis Dataset}
\label{sec:synthesis-dataset}

In this work, we broaden the scope of the dataset collection for scientific synthesis. The primary requirement for building a corpus of scientific syntheses is access to research problems or questions along with their associated papers. Previous efforts involved manually curating this information through a team of human annotators~\cite{core-gpt}. Here, we systematize this approach by utilizing the Open Research Knowledge Graph (ORKG), a crowdsourcing platform that provides structured research contributions and comparisons. The subsequent paragraphs detail our method for compiling a reliable collection of research problems and corresponding papers to generate a multidisciplinary scientific syntheses corpus. Our corpus surpasses previous work in both multidisciplinarity and size, containing three times as many data samples, as a direct consequence of relying on the ORKG crowdsourced data.

\subsection{Synthesis Generation Data Preparation}

\textbf{The ORKG data source.} The ORKG is a web-based service that structures scholarly research contributions into a knowledge graph, using a crowdsourcing approach where users add and semantically describe paper contributions (of which each paper may have one or more). A key feature, \textit{Comparisons}, allows users to select and compare multiple research contributions in a tabular format \citep{OlleenEtAlSciknow19}. These Comparisons, curated by users, include papers addressing specific research themes. For defining a corpus for scientific synthesis generation, we utilized ORKG Comparisons to extract human-annotated research problems and their associated papers, providing an ideal data source.

\textbf{Data processing.} The ORKG Python package (\url{https://pypi.org/project/orkg/}) was used to collect the data. First, we found all research problems with Comparisons, which produced 1,300 Comparisons for 708 research problems. From these Comparisons, we eliminated those with fewer than five unique papers, which left 495 Comparisons. The minimum threshold of five papers per comparison is a fixed criterion in this study for generating scientific syntheses. This criterion is based on popular search systems like \href{https://elicit.com/}{Elicit} or \href{https://ask.orkg.org/}{ORKG Ask}, which typically generate syntheses from the top five results of their respective search engines. Next, we sourced abstracts for each paper in the Comparisons using the open-access platforms \href{https://www.semanticscholar.org/}{Semantic Scholar}, \href{https://www.crossref.org/}{Crossref}, and \href{https://core.ac.uk/}{CORE}. This resulted in 329 Comparisons with five or more papers with abstracts. This step was essential since in this work, paper titles and abstracts are the context from which the syntheses are generated. Finally, for each Comparison, all papers were grouped as all possible collections of five contributions, with each collection representing one sample in the data, for a total of 541 samples. However, it is not uncommon for a Comparison to be tagged in multiple research fields, resulting in duplicated samples that differ only in research field. In such cases, only one Comparison was randomly selected to be kept, yielding a final dataset of 348 samples with each sample as a candidate for scientific synthesis generation.

\textbf{Intermediate ORKG scientific synthesis generation dataset.} The ORKG scientific synthesis generation dataset including the respective accompanying generated syntheses (details in \autoref{subsec:ssg}) is publicly released at \url{https://github.com/jd-coderepos/scisynthesis}. Each sample in the dataset consists of the following: sample ID, research field, research problem, and the title, abstract, and DOI (if available) of five papers. The research field is selected from the ORKG's own taxonomic schema (\url{https://orkg.org/fields}), with users permitted to choose any level within the hierarchy, whereas research problems are entered as free-form text. Due to the flexibility allowed in user inputs, the research problems in the ORKG vary widely in structure and specificity. For example, both ``text classification'' and ``Automated construction of health knowledge graphs from medical records'' are included as research problems in our dataset. However, the diverse and subjective nature of these entries, stemming from the dataset's crowdsourced origins, reflects and is representative of user behavior on online search platforms such as Elicit or ORKG Ask. This diversity is a specific strength of our dataset, setting it apart from previous datasets~\cite{core-gpt} created by a single team of human annotators. Similarly, since users may assign a research field from any level in the hierarchy, they range from broad, high-level fields like ``Chemistry'' to more specified fields like ``Medicinal Chemistry and Pharmaceuticals.''  Therefore, each sample has two research field columns: the original label selected by the user, and a mapping of the selected field to the third level of the ORKG's taxonomy. We chose the third level as representative enough to generalize to specific research field assignments. Note, research fields are not used in synthesis generation but are included as data points to help organize and understand the dataset. \autoref{tab:freq} lists the top 10 research fields in the ORKG synthesis generation dataset. The complete distribution is available at \url{https://github.com/jd-coderepos/scisynthesis/blob/main/corpus/domain_counts.xlsx}.

\begin{table}
\footnotesize
  \caption{Top 10 research fields in the ORKG synthesis dataset.}
  \label{tab:freq}
  \begin{tabular}{p{6cm}r}
    \toprule
\bf Research field& \bf Frequency\\ \midrule
Computer Sciences & 125\\
Physics & 28\\
Animal Sciences	& 19\\
Chemistry & 17\\
Urban Studies and Planning &16\\
Earth Sciences	&14\\
Oceanography and Atmospheric Sciences and Meteorology &14\\
Science and Technology Studies	&12\\
Materials Science and Engineering	&12\\
Engineering	&10\\
  \bottomrule
\end{tabular}
\end{table}

\subsection{Scientific Synthesis Generation}
\label{subsec:ssg}

\textbf{Task description.} Inspired by prior work~\cite{evans2024large}, the task of scientific synthesis is defined as a specialized form of multi-document summarization. It involves combining the main insights from multiple research papers---five in this work---into a coherent paragraph that addresses a specific research problem or question.

The scientific synthesis generation task encompasses the following five key characteristics. \textbf{1)} \textit{Use of scientific literature:} This process involves synthesizing information from the scientific literature, primarily from titles and abstracts of research papers. The task requires summarizing these texts and evaluating their relevance, correctness, and completeness concerning the research problem. \textbf{2)} \textit{Synthesis format:} 
The synthesis should be concisely presented in a single paragraph, limited to 200 words. This limit aligns with the standard guidelines for scientific abstracts as recommended by APA~\cite{Cherry2023}, MLA~\cite{wikiHowMLAAbstract}, Harvard~\cite{HarvardDissertation}, and Chicago~\cite{IEEEPVSCAbstract} style guides, aiming for brevity and precision in summarizing key information \cite{ScienceBuddiesAbstract, Nagda2013}. The format requires distilling and integrating diverse scientific insights into a coherent, comprehensive summary that addresses the research problem directly. This single-paragraph approach emphasizes the need for concise and cohesive communication of complex information.
\textbf{3)} \textit{Synthesize vs. summarize:} The objective is to synthesize—meaning to combine elements into a coherent whole—rather than merely summarizing each source individually. This involves the integration, cohesion, and coherence of information from multiple sources to produce new insights or understanding in response to the research problem. \textbf{4)} \textit{Referencing source material:} Each claim or piece of information in the synthesis must be traceable back to the source material (the abstracts) to ensure the synthesis's accuracy and reliability. \textbf{5)} \textit{Adherence to quality characteristics:} Building on approaches from linguistic and semantic quality evaluation studies of generated texts~\cite{fu2023gptscore}, the quality of the synthesis should satisfy the following nine key criteria: relevancy, correctness, completeness, informativeness, integration, cohesion, coherence, readability, and conciseness. These criteria, identified as apt for synthesis quality evaluation, collectively ensure effective communication of the synthesized information. The role and the application of these criteria in practice in this work are introduced in \autoref{subsec:ninecriteria}.

To accommodate the diverse information scope typically found in research papers, and to tailor the synthesis to the specific interests of the user, we developed three distinct types of synthesis: \textbf{1)} \textit{Paper-wise synthesis}, which provides a general overview; \textbf{2)} \textit{Methodological synthesis}, which concentrates on the methods and their pertinent details; and \textbf{3)} \textit{Thematic synthesis}, which focuses on identifying and summarizing recurring themes or patterns. Each synthesis type is designed to address a specific research problem while aligning with the intended focus of information. For example, a methodological synthesis will include only details about the methods discussed in the paper abstracts. In contrast, a thematic synthesis will specifically target the repetitive overarching themes present in the research.

As alluded to in the Introduction, the goal of this work is to develop a systematic framework for generating scientific syntheses using Large Language Models (LLMs4Synthesis). To our knowledge, this is the first framework to showcase the integration of both generation and evaluation components, aimed at optimizing the downstream synthesis model. While implemented in the context of ORKG Ask, LLMs4Synthesis is designed to be easily adaptable to other similar platforms like Elicit. 
The remainder of this paper discusses our use of generative AI technology, specifically LLMs, to achieve this goal. We start by describing the LLMs used for scientific synthesis generation and their application in the next paragraph.

\textbf{Synthesis generation task instruction, models, and the final ORKG syntheses dataset.} This task involved using an LLM to generate a synthesis from the titles and abstracts of five scientific papers, tailored to the synthesis type and research problem.

Previous studies on scientific synthesis generation~\cite{core-gpt,evans2024large}, have not extensively discussed how LLMs are prompted for this task. We designed the scientific synthesis generation prompt based on prompt engineering best practices~\cite{Saravia_Prompt_Engineering_Guide_2022} and various tested prompts. Our prompt for synthesis generation is detailed in \autoref{tab:synthgenprompt}. The prompt includes two main parts: a detailed task specification and placeholders for input papers. The first part specifies the task to the LLM, filling in the ``[research-problem]'' placeholder with the research problem from the ORKG synthesis dataset and ``[prompt-type-input-instruction]'' with instructions corresponding to the synthesis type. For example, the methodological synthesis type instruction is: ``The objective of this synthesis is to focus on the methodology. Therefore, compare and integrate the methodologies used in each paper content, emphasizing how they contribute to the research problem.'' The prompts for all three synthesis types are available at \url{https://github.com/jd-coderepos/scisynthesis/tree/main/synthesis-generation-prompts}. Based on initial feedback from the LLM's responses on a subset of the dataset, the first part of the prompt was iteratively refined to enhance specificity and instructional emphasis (notably, the 200-word limit is reiterated for clarity). The second part of the prompt includes placeholders for the titles and abstracts of five papers which are to be given as input to the LLM to perform the synthesis generation task.

\begin{table}[!t]
\caption{Scientific synthesis generation task prompt.}
\label{tab:synthgenprompt}
\footnotesize
\begin{tabularx}{\columnwidth}{|X|} \toprule
\textbf{Part I - Task Instruction: Synthesis Task Specification} \\ \hline
Generate a synthesis from the provided papers as content on the research problem "[research-problem]" into a concise single paragraph of no more than 200 words. Follow these instructions: \\
- Only the titles and abstracts from exactly five scientific papers will be provided, to be used as content for the synthesis. \\
- "[prompt-type-input-instruction]". \\
- Support each claim with citations, formatted as (1) or (3, 5) to refer to the respective papers' content. \\
- Ensure the output is a single cohesive paragraph without section headings, titles, abstracts, or any paper-like structure. \\
- Focus on essential information, maintaining clarity and precision. \\
- Do not include additional information or exceed the specified word count of 200 words. \\ \midrule
\textbf{Part II - Task Input: Five Papers to Synthesize} \\ \hline
\textbackslash{}n\textbackslash{}n Papers 
\textbackslash{}n 1. "[content-1]" 
\textbackslash{}n 2. "[content-2]" 
\textbackslash{}n 3. "[content-3]" 
\textbackslash{}n 4. "[content-4]" 
\textbackslash{}n 5. "[content-5]" \\ \bottomrule
\end{tabularx}
\end{table}

We applied two prominent state-of-the-art LLMs for the scientific synthesis generation task: the open-source Mistral-7B~\cite{jiang2023mistral7b}, and the proprietary GPT-4~\cite{achiam2023gpt4}. Thus the resulting \href{https://github.com/jd-coderepos/scisynthesis/tree/main/synthesis-generation-prompts}{ORKG syntheses dataset} comprises three syntheses data files, per synthesis type, per LLM (six files overall). Each file contains the raw synthesis response and the text of the synthesis. So total of 1044 syntheses for each model and an overall total of 2088 syntheses.





\section{Synthesis Quality Evaluation}
\label{sec:quality-eval}

To enhance our understanding of the weaknesses of LLMs concerning the synthesis objective and toward realizing potential improvements within the LLMs4Synthesis framework, we carried out a two-part qualitative evaluation of the generated syntheses. This evaluation included: 1) an LLM-based assessment using the most powerful variant of the GPT models, specifically GPT-4-Turbo, and 2) a human survey evaluation of a subsample from the synthesis dataset. Training moderate-sized open-source models effectively we demonstrate with Mistral 7B. We aim to make synthesis generation tasks more accessible and decrease dependence on proprietary models, which, while highly effective as indicated by popular \href{https://huggingface.co/spaces/lmsys/chatbot-arena-leaderboard}{LLM leaderboards}, often lack transparency in their pre-training methods, datasets, and parameters. This is a barrier to deeper research insights and wider community participation. Consequently, our work is dedicated to developing methods that equip open-source LLMs to handle essential tasks in scientific language processing.

As a first step, we established a comprehensive set of evaluation criteria for the generated syntheses, detailed in the next section.

\subsection{Nine Criteria of Synthesis Quality}
\label{subsec:ninecriteria}

Prior work~\cite{evans2024large} has provided a thorough analysis of the limitations associated with existing quantitative evaluation metrics, such as the \textsc{rouge} family \cite{lin-2004-rouge}, traditionally used for text summarization tasks—-a broader context for the synthesis objective. The limitations highlighted the necessity to establish evaluation criteria that focus on linguistic and semantic qualitative aspects. In this vein, inspired by a prior study that assessed synthesis quality using just three criteria—\textit{comprehensive}, \textit{trust}, and \textit{utility}—-considered crucial for high-quality synthesis \cite{core-gpt}, this work aims to expand on this foundation. We seek to develop a broader set of criteria for synthesis quality evaluation, incorporating thorough linguistic and semantic assessments, informed by insights from related research on text quality evaluation in summarization~\cite{fu2023gptscore}. After filtering out unrelated criteria from existing work, we identified nine criteria to evaluate the quality and effectiveness of the synthesized information. Each criterion is presented to the evaluator in the form of a question, as outlined below. 

\textbf{1. Relevancy:} Is the information in the answer relevant to the problem? \textbf{2. Correctness:} Is the information in the answer a correct representation of the content of the provided abstracts? \textbf{3. Completeness:} Is the answer a comprehensive encapsulation of the relevant information in the provided abstracts? \textbf{4. Informativeness:} Is the answer a useful and informative reply to the problem? \textbf{5. Integration:} Are the sources structurally and linguistically well-integrated, using appropriate markers of provenance/quotation and logical connectors for each reference? In addition, are the sources integrated within a single paragraph? \textbf{6. Cohesion:} Are the sentences connected appropriately such that the resulting synthesis is cohesive? \textbf{7. Coherence:} Are the ideas connected in a sound and logical manner? \textbf{8. Readability:} Does the answer follow appropriate style and structure conventions for academic writing and use language correctly? \textbf{9. Conciseness:} Is the answer short and clear, without redundant statements? Furthermore, is the synthesis approximately 200 words long?

Our initial five criteria were crafted to enhance objectivity and precision in evaluating syntheses, expanding on the three criteria used earlier in synthesis evaluation~\cite{core-gpt}. Specifically, our `3. completeness' corresponds to their `comprehensive,' our `2. correctness' and `5. integration' relate to their `trust,' and our `1. relevancy' and `4. informativeness' align with their `utility.' Thus, we expanded the three foundational criteria into five. Our remaining four criteria, three—`6. cohesion,' `7. coherence,' and `8. readability'—are rooted in established text evaluation practices for linguistic quality, which are, for instance, commonly applied in summarization scenarios. The final criterion, `9. conciseness,' is unique to this work. It emphasizes both the quality of the synthesized information and adherence to the specified word limit, essential for presenting a succinct representation of synthesized research insights to the reader.


\subsection{LLM Evaluation of Synthesis Quality}
\label{subsec:llmeval}


Prior work has demonstrated the effectiveness of LLMs as automatic evaluators for scientific synthesis~\cite{evans2024large,fu2023gptscore,bai2024benchmarking}, arguing that they are becoming both the tools and the standards for language assessment.
Thus, for the evaluation task, we utilized GPT-4-Turbo, the most advanced LLM available at the time of writing. The evaluation prompt and script are available at\url{https://github.com/jd-coderepos/scisynthesis/tree/main/gpt-4%20synthesis-evaluator}. The prompt includes all nine evaluation criteria along with their corresponding questions. Each evaluation criteria was assessed based on a rating scale from 1 to 5 as follows: 1. Very Bad, 2. Bad, 3. Moderate, 4. Good, and 5. Very Good. Each rating level was specifically tailored with a description to aid in the assessment of each criterion.
For instance, consider the rating scale description for the first criteria i.e. ``1. Relevancy: is the information in the answer relevant to the problem?'' as follows: \textbf{Rating 1. Very bad:} The information provided does not relate to the research problem, showing a lack of understanding or connection to the topic. \textbf{Rating 2. Bad:} The information occasionally relates to the research problem but lacks direct and consistent relevance. \textbf{Rating 3. Moderate:} The information is generally related to the research problem, with occasional lapses in direct relevance. \textbf{Rating 4. Good:} The information is consistently relevant to the research problem, with only minor exceptions.
\textbf{Rating 5. Very good:} The synthesis is directly and consistently relevant to the research problem, demonstrating a deep understanding of the topic and its nuances. For the full prompts, including detailed descriptions of the rating scale for all criteria, please refer to the \href{https://github.com/jd-coderepos/scisynthesis/blob/main/gpt-4%20synthesis-evaluator/evaluation-system-prompt}{system prompt}. Finally, similar to 3-fold cross-validation, the GPT-4 synthesis evaluator was run thrice on the same data samples to ensure consistent assessment. The resulting scores, which did not show significant divergence largely due to our detailed prompt specification of the expected evaluation task leaving little room for ambiguity in interpretation, were averaged and reported.

\subsubsection{Results.} 
\begin{figure}[h]
  \centering
  \includegraphics[width=\linewidth]{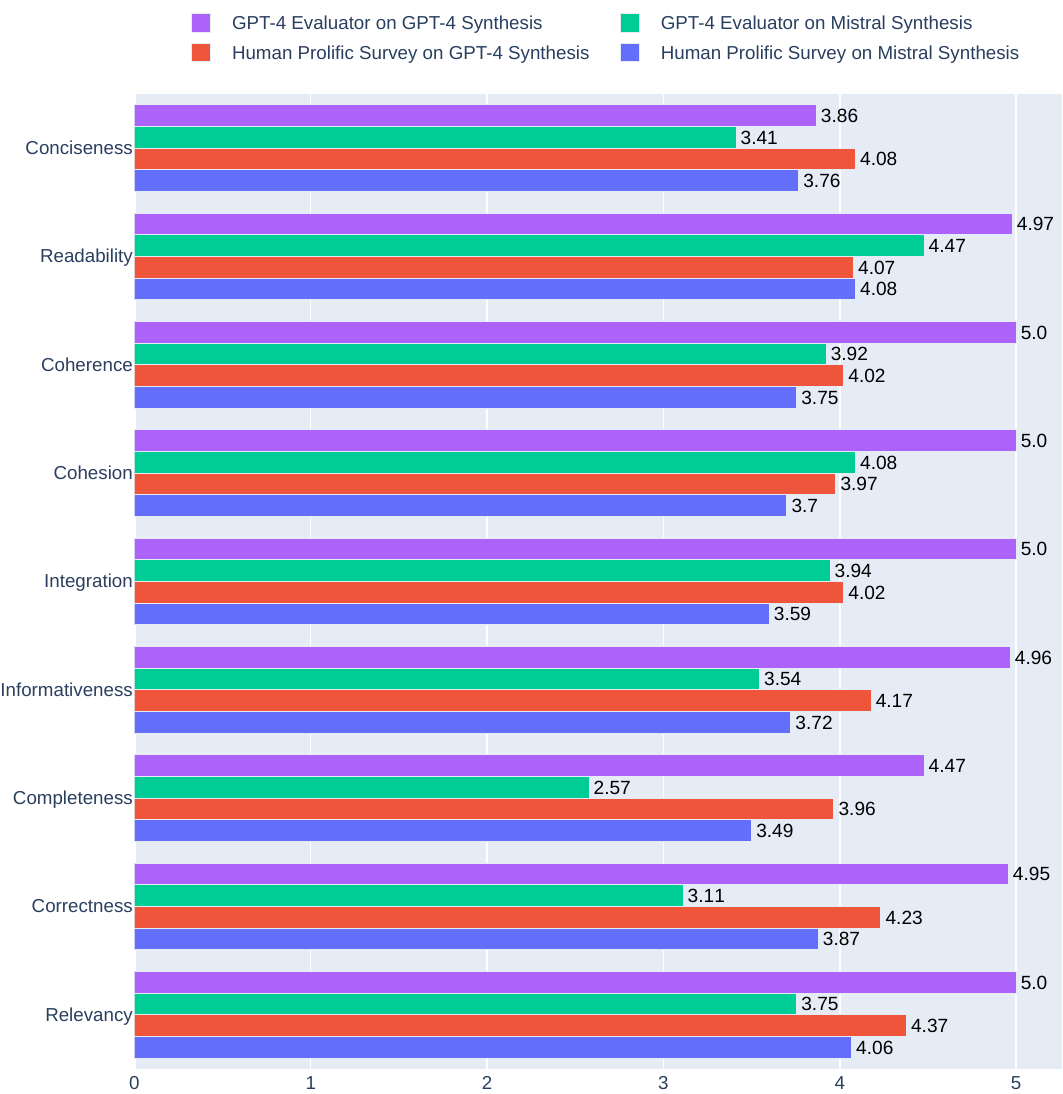}
  \caption{Evaluation results from the GPT-4 LLM evaluator (purple and green bars) and a Prolific human survey (red and blue bars) for syntheses generated by Mistral and GPT-4. The data includes averaged scores across three synthesis types and five domains—Chemistry, Computer Science, Earth Science, Linguistics, and Sociology. 
  }
  \label{fig:subsampleeval}
\end{figure}

The results are depicted as the purple and green bars in \autoref{fig:subsampleeval}. These results per synthesis quality criteria are averaged for the three synthesis types, i.e. paper-wise, methodological, and thematic since the differences between their scores were observed to be minimal. Comparing the two synthesis generators, GPT-4 outperforms Mistral on all characteristics. On the GPT-4-generated syntheses, the highest-scored characteristics are integration, cohesion, coherence, and readability, all of which have average scores of 4.95 or greater, reflecting strong overall structure, logical flow, and ease of comprehension. Informativeness, correctness, and relevancy were also scored highly as indicators of the trust and utility of the answers. Furthermore, the evaluator was conservative in scoring completeness and conciseness suggesting that there might be some redundancy or verbosity in the syntheses. Mistral-7B was evaluated substantially lower than GPT-4. Notably, Mistral and GPT-4 differ significantly in size, with GPT-4 rumored to exceed a billion parameters, which likely contributes to its strong performance in text generation and synthesis tasks. Completeness, correctness, and conciseness received the lowest scores, attributed to the unexpected output format of the syntheses produced by Mistral. In these cases, syntheses were wrongly structured in the shape of original research articles comprising title, abstract, keywords, introduction, results, methodology, and conclusion rather than as a paragraph-format synthesis of multiple other works.


In conclusion, both models performed best on readability, integration, cohesion, and coherence, demonstrating the fitness of LLMs to generate clear, logically structured texts. However, there may be a difference in the suitability of each LLM to the varying synthesis types, with GPT-4 performing slightly better on methodological, followed closely by thematic, whereas Mistral performs best on thematic, followed by paper-wise.

\subsection{Human Evaluation of Synthesis Quality}

We recruited human evaluators to conduct a survey assessing synthesis quality. 

\subsubsection{Survey setup} Prolific (\url{https://www.prolific.com/}), a paid crowdworker platform specifically for academic studies, was used to conduct the survey. To reduce effort, we subsampled the syntheses dataset to include a subset of syntheses from the domains of Chemistry, Computer Science, Earth Science, Linguistics, and Sociology. These domains were selected because from surface observations the Mistral syntheses received a wide range of scores by the LLM evaluator (cf. \autoref{subsec:llmeval}), and represent scholarly diversity. For each domain, two data samples were selected, each including a research problem, sets of five papers, and six generated syntheses-—three each from Mistral-7B and GPT-4, corresponding to the three different synthesis types. The samples were chosen to exhibit distinct performance levels: one with high average evaluation scores and the other with low scores from the LLM evaluator for the Mistral syntheses. This approach aimed to compare the scoring variability between human and LLM evaluators. Furthermore, Mistral, our open-source LLM, is targeted for performance enhancement in synthesis generation through the LLMs4Synthesis framework, which justifies our focus on its generated syntheses' evaluation scores.

Five surveys were created for each of the five domains, with each survey including both data samples thus a total of 12 syntheses. Each survey started with an introduction screen that defined the task as well as detailed the nine evaluation criteria and the 5-rating scale. In a subsequent screen, participants were shown the research problem and paper titles and abstracts before evaluating the six LLM-generated syntheses on sequential screens. They could revisit the abstracts at any time and optionally provide free text feedback for each characteristic. Links to the original surveys underlying the domain names are: \href{https://forms.gle/xCTwbv14aqFYCEMx6}{Chemistry}, \href{https://forms.gle/9vinWXPZ7BdA8hRg8}{Computer Science}, \href{https://forms.gle/rueqNwvpj9RSRYUP6}{Earth Science}, \href{https://forms.gle/HYcRUKCP2Vuzover9}{Linguistics}, \href{https://forms.gle/NGMN64citAmsT3SF9}{Sociology}. Evaluations were obtained from three participants for each domain, involving 15 participants in total. The survey for the human evaluator and the \href{https://github.com/jd-coderepos/scisynthesis/blob/main/gpt-4%20synthesis-evaluator/evaluation-system-prompt}{LLM evaluator prompt} were closely aligned to ensure nearly identical evaluation setups.

\subsubsection{Survey participant characteristics.} Prolific filters were used to screen for eligible individuals, specifically: 1) those fluent in English; 2) those who have completed at least an undergraduate degree, and; 3) those whose subject is in the appropriate domain. Prolific only verifies participants' identity and country of residence; all other demographic details, such as language proficiency and academic specialization, are self-reported.

Our survey participants, aged 22 to 33, included both undergraduate and graduate students. Detailed demographic information is available \href{https://github.com/jd-coderepos/scisynthesis/tree/main/corpus/prolific}{online}. Participants were paid \pounds9 (GBP) per hour, with an estimated survey duration of 2 hours. 

\subsubsection{Survey Results.} The results are depicted as the red and blue bars in \autoref{fig:subsampleeval}. In line with the LLM evaluator results, overall, GPT-4-generated syntheses consistently outperform those from Mistral across all evaluated characteristics, although some differences are slight. The scores from both LLM and human evaluators are based on the same data subsample used for the Prolific survey. Performance across the evaluation criteria varied. In the case of readability, both generators performed comparably; however, this was Mistral's strongest characteristic while for GPT-4 the average scores for relevancy, correctness, informativeness, and conciseness exceeded that of readability.  The most significant difference in model performance is in the characteristics of completeness, informativeness, and integration. Additionally, we can gain an overview of participants' perceptions of the syntheses by examining the optional feedback provided for some texts.  Below is a summary of the comments left for each characteristic.

\textbf{1. Relevancy.} Both LLMs are generally positively received for relevancy. GPT-4's syntheses are more detailed and directly relevant to research problems, while Mistral may generalize findings. Both models need to improve on making connections and providing deeper insights, but GPT-4 slightly outperforms Mistral. \textbf{2. Correctness.} GPT-4 shows a high degree of correctness, whereas Mistral's performance is variable, with reports of ``multiple serious inaccuracies.'' While GPT-4 reliably captures abstract details, it occasionally misses specific study details or statistical values. Mistral sometimes mislabels studies and includes incorrect information. \textbf{3. Completeness.} GPT-4 generally covers abstracts better and includes more specific information than Mistral, although both can miss key quantitative details. Mistral has more frequent gaps and occasionally misses entire studies, often being too brief. Both need to enhance their depth and detail, but GPT-4 maintains slightly better completeness. \textbf{4. Informativeness.} GPT-4 is generally more informative, though more explicit details on methods and influencing factors would be beneficial. Mistral's syntheses tend to be general and lack depth, with both models critiqued for insufficient insight into the research problem. \textbf{5. Integration.} GPT-4 effectively integrates sources into a structured synthesis, whereas Mistral tends to list sources without transitions. GPT-4 is more consistent in creating a unified narrative, though both could improve paragraph organization. \textbf{6. Cohesion.} GPT-4 offers more cohesive syntheses with smoother transitions and well-connected ideas, whereas Mistral's work often reads like separate summaries. GPT-4 effectively groups studies thematically, enhancing synthesis cohesion. \textbf{7. Coherence.} GPT-4 shows higher overall coherence, while Mistral's syntheses can seem disjointed. Both models could strengthen connections between ideas, but GPT-4 is closer to achieving this. \textbf{8. Readability.} Both models produce clear syntheses, but Mistral's style is slightly preferred for its simplicity. GPT-4 adheres well to stylistic norms but can be overly complex. Despite structural issues, Mistral's writing quality is praised. \textbf{9. Conciseness.} GPT-4 generally adheres to word limits better, while Mistral often exceeds or falls short of the 200-word target. Both are praised for being non-repetitive.

Finally, the domain-specific survey evaluations reveal that GPT-4 performs best in Earth Science and Computer Science, while Mistral excels in Earth Science and Chemistry. Both LLMs show weaker performance in Linguistics and Sociology, suggesting a better fit for engineering and hard sciences. For more insights, visit our \href{https://github.com/jd-coderepos/scisynthesis}{repository}.

This first part of the paper has introduced the scientific synthesis task, LLM synthesis generators, the multidisciplinary ORKG dataset, and insights from both an automatic LLM evaluator and a human survey. These elements contribute to defining the LLMs4Synthesis framework, which optimizes open-source LLMs for synthesis generation. The second part evaluates this framework, particularly against Mistral-7B, focusing on improvements in output format and addressing weaknesses in the nine evaluation criteria.

\section{The LLMs4Synthesis Framework}
\label{sec:llms4synth}

\begin{figure*}[!htb]
  \centering
  \includegraphics[width=\linewidth]{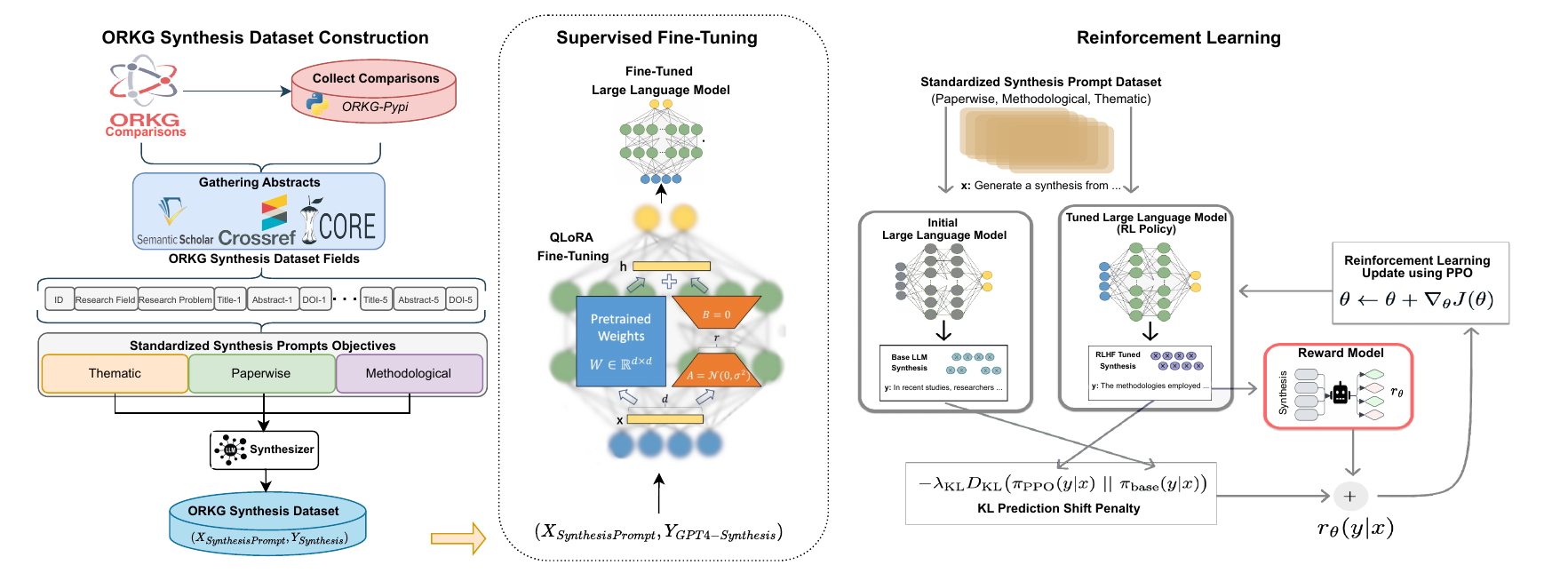}
  \caption{LLMs4Synthesis Framework using Supervised Fine-Tuning and Reinforcement Learning~\cite{lambert2022illustrating}. Note: SFT is optional, but we achieved better performance when it was included.}
  \label{LLMs4Synthesis}
\end{figure*}
%
The proposed LLMs4Synthesis adopts a reinforcement learning with AI feedback (RLAIF)~\cite{christiano2017deep,bai2022constitutional} paradigm to fine-tune LLM synthesizers for scientific synthesis generation. As illustrated in~\autoref{LLMs4Synthesis}, the first module attempts to construct a standardized dataset. The next step of LLMs4Synthesis is supervised fine-tuning with the aim of learning the task distributions by instructing Mistral-7B as the base LLM and GPT-4 synthesis as the gold standard, attempting to shift the general LLM probability distributions toward scientific domains while maintaining the general knowledge. Finally, LLMs4Synthesis uses reinforcement learning (RL) with Proximal Policy Optimization (PPO)~\cite{schulman2017proximalpolicyoptimizationalgorithms} algorithm to leverage human/AI feedback to guide the model's learning process, aiming to enhance the model's performance on synthesis generation by aligning its outputs with human preferences.

\subsection{Supervised Finetuning (SFT)}
Supervised fine-tuning (SFT) of LLMs~\cite{minaee2024largelanguagemodelssurvey} for scientific synthesis involves adapting the model to generate synthesized content from input papers using labeled training data. This process is crucial for customizing a pre-trained model, originally trained on a diverse range of texts, to handle more specific tasks like scientific synthesis generation based on provided research papers. The SFT module in the LLMs4Synthesis framework, illustrated in \autoref{LLMs4Synthesis}, uses a standardized prompt template that incorporates abstracts and titles from five given papers as input $X_{SynthesissPrompt}$, with the synthesis generated by GPT-4 serving as the ground truth $Y_{GPT4-Synthesis}$.

We used a Quantized Low-Rank Adapter (QLoRA)~\cite{dettmers2023qloraefficientfinetuningquantized} for SFT to adapt LLM to a synthesis generation task while minimizing computational resources and memory usage. Let's consider $M(W)$ as a pre-trained Mistral-7B LLM, where $W\in\mathbb{R}^{d*d}$ represents the model's parameters, where $d$ is the dimension of the LLM hidden state. Adapter layers with low-rank parameterization~\cite{hu2021loralowrankadaptationlarge} introduce the transformation based on $W^{\prime} = A B^T$, where $A\in \mathbb{R}^{d*r}$ and $B\in \mathbb{R}^{d*r}$ are low-rank matrices with $r$ representative of the rank, which set to $r=8$. $A$ is the transformation of the hidden state of LLM, $B$ is the further transform the output of $A$. Later, $A$ and $B$ quantized via $A_q = quant (A)$ and $B_q = quant(B)$ to reduce the memory footprint using 4-bit quantitation at $quant(.)$. In forward pass generation the model output $h$ is been obtained based on input sequence $x \in X_{SynthesissPrompt}$ by $h = M(W, x) + Dropout(\alpha *A_q(B_{q}^{T} M(W, x)))$, Where $ M(W, x)$ is the output of $M$ for $x \in X_{SynthesissPrompt}$, $A_q(B_{q}^{T} M(W, x))$ is the task-specific adaptation introduced by quantized adapter layer, $\alpha$ is the scaling factor for balancing the contribution the model final output which $\alpha=16$, and $Dropout(.)$ denotes the dropout operation which applied to the output of the low-rank adapter with dropout rate of $0.05$ to regularize the low-rank adapters, reducing overfitting by randomly dropping parts of the adapter’s output during training.

Later, the \textit{AdamW} optimizer~\cite{loshchilov2019decoupledweightdecayregularization} was employed with a learning rate of $2*10^{-4}$, a gradient accumulation step of $4$, and a warmup step of $0.03$. This configuration was used to optimize both the LLM parameters $M(W)$ and the low-rank adapter parameters $A_q$ and $B_q$ over $5$ epochs.

\subsection{Modeling Feedback}
Incorporating desired alignment preferences through reward modeling is essential for enhancing the performance of language models. This section presents two reward functions aimed at guiding the generation of syntheses. The first approach employs basic features to align with synthesis format as a paragraph and word limit constraints as structural preferences using feedback, while the second utilizes detailed qualitative scoring to encourage high-quality output using AI feedback, reflecting user objectives and enhancing alignment with human values by using GPT-4 as a proxy to the reward function~\cite{kwon2023rewarddesignlanguagemodels}.
\subsubsection{Basic Features} \label{humanpref}
In synthesizing scientific content using LLMs, a key challenge is ensuring that the generated synthesis aligns with desired format and constraints. As noted in the earlier evaluations, Mistral produced synthesis following a conventional paper structure which is not per our specified synthesis format i.e. as a paragraph and within 200 words. 
To address this, a rule-based paper structure identifier was developed with two goals: (1) to discourage overly structured synthesis, and (2) to promote succinctness. We found that 35.44\% of the Mistral generated synthesis dataset exhibited a paper structure, detected through 17 specific academic terms (e.g. "Title", "Abstract", "Conclusion", etc.) and fabricated author names in the citations. To address this, we used nine regular expressions to identify various reference forms, helping us quantify the adherence to academic styles in LLM-generated texts. The word count limit used a standard space-based splitting method.  

Let's consider $S$ as a synthesis, $WC$ as a word counter function, and $PS$ as a paper structure identifier function that returns 1 if the synthesis has the paper structure and 0 otherwise. The basic features try to adapt human preferences for obtaining an ideal structural format of synthesis $S$ by the following reward function:
\[
R_{basic}(S) =
\begin{cases}
-1.5 & \text{if } WC(S) < 50 \\
-1 & \text{if } WC(S) > 200 \\
-2 & \text{if } PS(S) = 1 \\
-0.5 & \text{if } \left| WC(S) - 200 \right| \leq 20 \\
2 & \text{otherwise}
\end{cases}
\]
The overall objective of this reward function $R_{basic}(S)$ is to encourage syntheses that are of moderate length (close to 200 words) and do not adhere strictly to a rigid paper structure. It penalizes both very short and very long syntheses, as well as those that follow a specific structure while rewarding those that meet the criteria for ideal synthesis length and flexibility. This approach reflects a balance between length, structure, and flexibility, aligning with human preferences for synthesis quality.

\subsubsection{GPT-4 Features}
To reward the quality of synthesis based on nine qualitative criteria, we introduce a reward function designed to reflect how closely the provided synthesis is aligned with ideal qualitative standards obtained by GPT-4. We defined the $PVF$ function that asses how close each score is per criteria to the preferred value, which in our case is five (the optimal value that one synthesis could get). The function is given by:
\[
    PVF_{score}(C, pv) = -\frac{1}{n} \sum_{i=1}^{n} \left| C_i - pv \right|
\]
Here $C$ represents a list of scores for nine qualitative criteria, denoted as $\{C_1, C_2 ..., C_n\}$ (where $n=9$), with each score $C_i$ within the range of $[1, 5]$. This reward aims to maximize the average score of $C$. To achieve this, we set the preferred value of $pv=5$. By using $PVF_{score}(C, pv)$ function, we aim to encourage LLM to produce a synthesis with higher overall scores based on qualitative criteria. Finally, the following reward function is derived, which rewards the synthesis based on the synthesis scores $C$ using the GPT-4 Evaluator.

 \[
  R_{GPT-4}(C) =
  \begin{cases} 
  4.0 & \text{if } PVF_{score} \geq -0.125 \\
  PVF_{score} & \text{otherwise}
  \end{cases}
  \]
  
Withing this reward function $R_{GPT-4}(C)$, we aim to provide positive rewards for values of ${PVF_{score}}$ that are closer to zero, as this indicates scores that are near the preferred value. To facilitate this, we set a threshold of $-0.125$. If the computed ${PVF_{score}}$ is greater than or equal to this threshold, which means it is closer to zero and therefore represents better alignment with the preferred value, the reward is adjusted to a higher fixed value of 4.0. Otherwise, the ${PVF_{score}}$ will be used as a reward score to punish the LLM for better synthesis generation.

\subsection{Reinforcement Learning}
To train a policy that generates higher-quality synthesis, we utilized reward models within an RLAIF framework. AI is utilized as a substitute for the traditional human feedback typically employed in similar contexts. In RL, given the current state (standardized prompts from the ORKG synthesis dataset), the LLM as a synthesizer produces an action (synthesis), that modifies the current state by adding a token to the current sequence. Once a full textual sequence has been produced, we can obtain a reward by rating the quality of the synthesis using reward models to mimic human preferences. To fine-tune the model with RL, we simply alternate between collecting data from the environment — done by generating text with the LLM and then scoring it with the reward model — and updating the policy according to the Proximal Policy Optimization (PPO) algorithm~\cite{schulman2017proximalpolicyoptimizationalgorithms}.

Importantly, we incorporate a term in the reward function that penalizes the Kullback-Leibler (KL) divergence~\cite{Joyce2011} by measuring the entropy of the token probability distributions learned by the RL policy $\pi_{PPO}(y|x)$, where this policy is derived from the tuned LLM synthesizer, compared to the initial LLM synthesis $\pi_{base}(y|x)$ from the frozen SFT synthesis. The final reward function will be calculated as follows:
$$R = r_{\Theta}(y|x) - \lambda DL_{KL}(\pi_{PPO}(y|x) || \pi_{base}(y|x)) $$

Where $r_{\Theta}(y|x)$ represents the reward model, which can be either $R_{basic}$ or $R_{GPT-4}$. The KL penalty coefficient parameter, denoted as $\lambda$, is set to $0.2$ for adaptive KL divergence control. We employed the \textit{Adam} optimizer with a learning rate of $2.94 \times 10^{-5}$ in conjunction with PPO policy, which involves $10$ epochs per batch of samples. The $DL_{KL}$~\cite{stiennon2022learningsummarizehumanfeedback} is defined as follows:
$$DL_{KL}(\pi_{\text{PPO}} \| \pi_{\text{base}}) = \log \left[ \frac{\pi_{\text{PPO}}(y|x)}{\pi_{\text{base}}(y|x)} \right]$$

Intuitively, the entropy value captures how much information is stored within a probability distribution by measuring the dissimilarity between two probability distributions, with the primary goal of maximizing the reward achieved by the model during RL policy training. Another advantage of KL divergence is that it constrains the model to ensure that it doesn't drift too far from the pre-trained policy.

\section{The LLMs4Synthesis Evaluations}
\label{sec:llms4syntheval}

\begin{table*}
\caption{Results from LLMs4Synthesis using both \textit{Basic} and \textit{GPT-4} evaluations across various metrics. \textsc{Word Count} metrics are reported as percentages, except for the average value. \textsc{Paper Structure} measures the percentage of observed synthesis with the paper's structure. \textsc{GPT-4} results are averaged scores across nine characteristics, obtained from three separate evaluations.}
\label{tab:results}
\footnotesize
\begin{tabular}{|l|l|l|r|r|r|r|r|r|r|}
\hline
\multicolumn{3}{|c}{\textbf{Evaluation Criteria}} & \multicolumn{3}{|c|}{\textbf{Systems}} & \multicolumn{2}{|c}{\textbf{w/ Basic Features}} & \multicolumn{2}{|c|}{\textbf{w/ GPT-4 Features}} \\
\cline{1-10}
Type & \multicolumn{2}{l|}{Metrics} & GPT-4 & Vanilla & SFT & RL  & SFT +  RL  & RLAIF & SFT + RLAIF \\
\cline{1-10}
\multirow{6}{*}{\textit{Basic}}  & \multirow{5}{*}{\textsc{Word Count}}  
                        & $WC<50$                &  \textbf{0.0} & \textbf{0.0} &  \textbf{0.0} & \textbf{0.0} & \textbf{0.0} &  0.42   &  \textbf{0.0}  \\
                        & & $50\leq WC < 150$    & \textbf{0.0} & 17.09 & 0.42 & 15.81 & 0.85 & 11.96 & \textbf{0.0} \\
                        & & $150\leq WC\leq 250$ & 91.02 & 48.71 &  81.62 & 83.76 &  98.71 &  87.17   &  \textbf{100}  \\
                        & & $ WC > 250$          & 8.97 & 34.18 &  17.94 & 0.42 &  0.42 &  0.42   &  \textbf{0.0}  \\
                        & & \textsc{Average}     & 218 & 242 &  231 & 189 &  \textbf{204} &  194  &  \textbf{205}  \\
                        \cline{2-10}
                        & \multicolumn{2}{l|}{\textsc{Paper Structure}} & 1.28 & 32.47 &  1.28 & 1.70 &  \textbf{0.0} &  0.85   &  \textbf{0.0}  \\
\hline
\multirow{9}{*}{\textit{GPT-4} } & \multicolumn{2}{l|}{\textsc{Relevancy}} & \textbf{4.97} & 4.33 & 4.65  & 3.97 &  4.48 &  4.75  & \textbf{4.88}   \\
                        & \multicolumn{2}{l|}{\textsc{Correctness}}        & \textbf{4.93} & 3.66 & 3.63  & 2.97 &  3.09 &  4.39  & \textbf{4.75}  \\  
                        & \multicolumn{2}{l|}{\textsc{Completeness}}       & \textbf{4.42} & 3.00 & 3.13  & 2.26 &  2.55 &  3.54  & \textbf{4.19}  \\  
                        & \multicolumn{2}{l|}{\textsc{Informativeness}}    & \textbf{4.95} & 3.94 & 4.21  & 3.41 &  3.84 & 4.28   & \textbf{4.83}   \\  
                        & \multicolumn{2}{l|}{\textsc{Integration}}        & \textbf{4.99} & 4.41 & 4.73  & 3.67 &  4.54 & 4.74   & \textbf{4.89}   \\  
                        & \multicolumn{2}{l|}{\textsc{Cohesion}}	       & \textbf{4.99} & 4.44 & 4.73  & 3.81 &  4.51 & 4.78   & \textbf{4.89}   \\  
                        & \multicolumn{2}{l|}{\textsc{Coherence}}	       & \textbf{4.98} & 4.38 & 4.69  & 3.76 &  4.47 & 4.73   & \textbf{4.85}   \\  
                        & \multicolumn{2}{l|}{\textsc{Readability}}        & \textbf{4.96} & 4.59 & 4.43  & 4.33 &  4.16 & \textbf{4.90}   & 4.78   \\  
                        & \multicolumn{2}{l|}{\textsc{Conciseness}}        & \textbf{3.89} & 3.42 & 3.32  & 3.11 &  3.08 & \textbf{3.65}   & 3.64   \\  
\hline
\end{tabular}
\end{table*}

\subsection{Experimental Setup}

\subsubsection{Dataset Split}
The ORKG synthesis dataset is divided into training and testing splits to facilitate both training and evaluation processes. The overall dataset consists of 348 comparisons and 1044 standardized synthesis prompts. We partitioned the dataset by domain, allocating 20\% of the comparisons for testing and the remaining 80\% for training. This resulted in 78 comparisons with 234 standardized synthesis prompts for testing and 270 comparisons with 810 standardized synthesis prompts for training. For training purposes, the data is further divided into two equal parts: 135 comparisons and 405 synthesis prompts each for training LLMs (denoted as \texttt{Train-LLM}) and RL (denoted as \texttt{Train-RL}). Additionally, a small subset from the test set, comprising 10 comparisons and 30 standardized synthesis prompts, is used for Prolific human evaluations. This structured division ensures a comprehensive approach to model training and evaluation, leveraging diverse methodologies and human feedback.

\subsubsection{Experimental Models} 
In our experiments, we utilized a variety of models to assess the effectiveness of different fine-tuning and RL strategies. We used seven models as follows: \textbf{(1) GPT-4} is a state-of-the-art LLM developed by OpenAI. For our experiments, we used the \texttt{gpt-4-1106-preview} version. \textbf{(2) Vanilla} is a Mistral-7B model that has not undergone any fine-tuning, representing the raw capabilities of the Mistral-7B LLM. For our experiments, we utilized the \texttt{Mistral-7B-Instruct-v0.1} version. \textbf{(3) SFT} is a fine-tuned version of the \texttt{Vanilla model}. The fine-tuning was performed using the QLoRA method, with the GPT-4 based synthesis outputs serving as the ground truth text. We used the \texttt{Train-LLM} split from the training data for fine-tuning. \textbf{(4) RL (w/ Basic Features)} involves fine-tuning the \texttt{Vanilla model} using RL with a basic reward function $R_{basic}(S)$. \textbf{(5)  SFT+RL (w/ Basic Features)} is created by further fine-tuning the \texttt{SFT model} using RL with a basic features reward function $R_{basic}(S)$ that combines SFT and RL to enhance the performance w.r.t human preferences. \textbf{(6) RLAIF (w/ GPT-4 Features)} is the result of further fine-tuning the \texttt{RLAIF (w/ Basic Features) model} using a GPT-4 features reward function $R_{GPT-4}(C)$. \textbf{(7) SFT+RLAIF (w/ GPT-4 Features)} represents the final stage of our experiments, where the \texttt{SFT+RLAIF (w/ Basic Features) model} is further fine-tuned using RLAIF with GPT-4 features reward function $R_{GPT-4}(C)$.

\subsection{Results}
To study the research questions (RQs) of this work, we examine the results reported for the experimental models in~\autoref{tab:results}. This analysis will provide insights into how each model performs across different evaluation metrics and criteria.

\textit{\textbf{RQ1: What features are essential for qualitatively evaluating the informativeness and meaningfulness of a scientific synthesis?}} Given the results in~~\autoref{tab:results}, we analyze this question using \textit{GPT-4} evaluation metrics and comparison of results for \textit{w/ GPT-4 Features} based models with \textit{Systems} models.

\noindent\textbf{Assessing semantic quality. } \textit{GPT-4} and \textit{SFT + RLAIF (w/ GPT-4 Features)} excel in several key areas. They lead with the highest scores in relevancy (4.97 and 4.88), demonstrating their ability to produce highly relevant synthesis content. These models also score highest in correctness (4.93 and 4.75), reflecting their strong performance in generating accurate information. In terms of completeness, \textit{GPT-4} and \textit{SFT + RLAIF (w/ GPT-4 Features)} again lead with scores of 4.42 and 4.19, indicating more comprehensive topic coverage. Moreover, \textit{GPT-4} and \textit{SFT + RLAIF (w/ GPT-4 Features)} consistently perform well across additional criteria such as informativeness, integration, cohesion, coherence, readability, and conciseness, showcasing their superior ability to produce well-rounded and high-quality outputs. In contrast, \textit{Vanilla} and \textit{SFT} show lower scores in these areas, suggesting they are less effective in relevancy, correctness, completeness, and overall content quality.

\noindent\textbf{Impact of adding w/ GPT-4 Features-based feedbacks. } Incorporating GPT-4 Features-based feedback into the \textit{RLAIF} and \textit{SFT+RLAIF} models significantly enhances its performance across nearly all metrics compared to the baseline \textit{Vanilla} and \textit{SFT} models. This improvement indicates that the addition of GPT-4 Features feedback results in more semantically rich and coherent syntheses. By leveraging the $R_{GPT-4}(C)$ reward function, the model effectively aligns with human preferences and achieves superior quality and informativeness through RLAIF-based fine-tuning.

\textit{\textbf{RQ2: How can LLMs be fine-tuned to adhere to specific formatting standards like word limits and paragraph structure in scientific syntheses?}}
We address this question by analyzing the findings presented in~\autoref{tab:results} using results for \textit{Basic} evaluation metrics and comparison between \textit{w/ Basic Features} based models with \textit{Systems} models.

\noindent\textbf{Adherence to the word limit. }\textit{GPT-4} with an average word count of 218 obtains fairly close to the ideal of 200 word limit. However, the \textit{Vanilla} model with a higher average word count of 242, shows less control over the adherence to the word limit, and even supervised fine-tuning (\textit{SFT} averaged word count of 231) did not show an improvement. Nevertheless, \textit{GPT-4} majority of outputs are within the 150-250 word range (91.02\%), however, \textit{Vanilla} (48.71\%) and \textit{SFT} (81.62\%) fail to surpass this performance.  When it comes to minimizing overgenerating ($WC >250$), \textit{GPT-4} performs exceptionally well with a score of 8.97\%. \textit{SFT} also shows impressive performance, scoring 17.94\%, which is twice as good as the \textit{Vanilla} model's score of 34.18\%. Similarly, in generating less content ($50 \leq WC < 150$), the \textit{Vanilla} model falls short with an average score of 17.09\%. Using the $R_{basic}(S)$ reward function, both the \textit{RL (w/ Basic Features)} and \textit{SFT + RL (w/ Basic Features)} models produce word counts close to the ideal average. \textit{RL (w/ Basic Features)} averages 189 words, while \textit{SFT + RL (w/ Basic Features)} averages 204 words. \textit{SFT + RL (w/ Basic Features)} achieves 98.71\% within the 150-250 word range and has minimal overgeneration (0.42\%). This indicates that \textit{SFT + RL (w/ Basic Features)} adheres better to word limits due to fine-tuning with reward modeling that mimics human preferences.

\noindent\textbf{Adherence to the single-paragraph style. }The \textit{Vanilla} model has a significantly higher percentage (32.47\%), indicating a tendency to deviate from single-paragraph style synthesis and follow the paper structure. In contrast, \textit{GPT-4} shows the lowest average at 1.28\%, adhering well to style criteria.  Fine-tuning has been effective, with \textit{SFT} at 1.28\% and \textit{RL (w/ Basic Features)} at 1.70\%, demonstrating an approximately 31\% drop in following the paper structure compared to the \textit{Vanilla} model. Additionally, \textit{SFT+RL (w/ Basic Features)} excels in maintaining a single-paragraph style by closely imitating human preferences.


\textit{\textbf{RQ3: How can the balance be achieved in LLMs between maintaining formatting standards and ensuring semantically informative and meaningful scientific syntheses?}}
According to the \textit{SFT+RLAIF (w/ GPT-4 Features)} results in~\autoref{tab:results}, balancing the basic characteristics and semantic quality in LLM scientific synthesis involves leveraging advanced fine-tuning methods such as RLAIF with specialized reward functions such as $R_{basic}(S)$ and $R_{GPT-4}(C)$ accordingly. The \textit{SFT+RLAIF (w/ GPT-4 Features)} model demonstrates an effective balance, excelling in both basic and qualitative metrics. It maintains optimal word count distribution and avoids paper-like structures while achieving high scores in all qualitative criteria. This balance is likely achieved through iterative fine-tuning that focuses on both structural adherence and semantic quality, indicating that an integrated approach combining SFT and RLAIF with model feedback can effectively balance these aspects.

\section{Discussion}
\label{sec:discuss}

\begin{figure}[t]
  \centering
  \includegraphics[width=\linewidth, height=7cm]{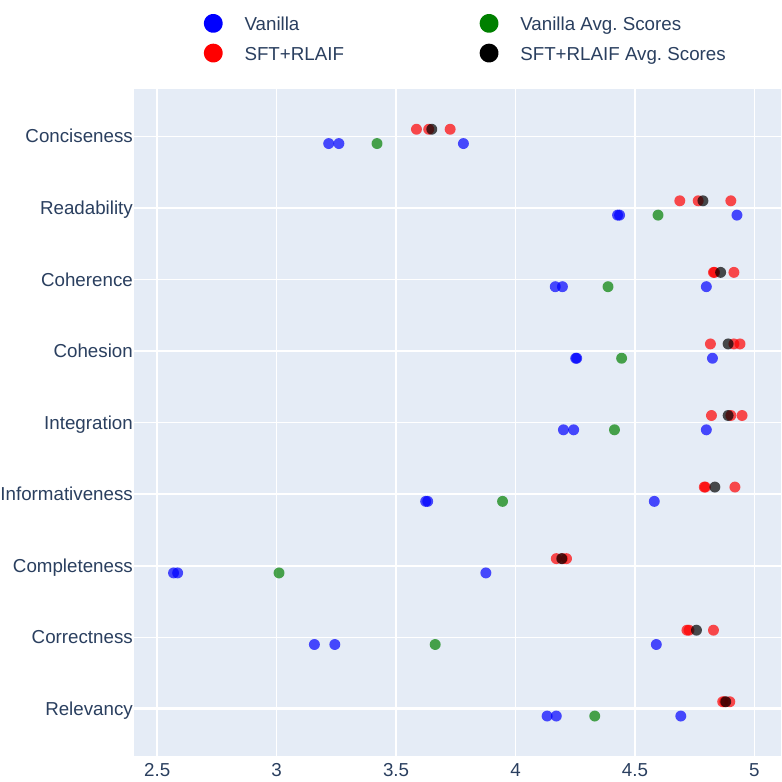}
  \caption{Consistency comparison of the GPT-4 evaluator between the \textit{Vanilla} and \textit{SFT+RLAIF (w/ GPT-4 Features)} models, assessed through three evaluations on the test set.}
   \label{consistency}
\end{figure}
\noindent\textbf{An LLM as a Quality Evaluator.} As language models like BERT and BART have advanced, evaluation methods have evolved. Traditional metrics have been supplemented by approaches such as BERTScore \cite{zhangbertscore}, which uses contextual embeddings to assess semantic similarity effectively, and BLEURT \cite{sellam2020bleurt}, which leverages human-annotated data for more accurate text quality predictions. The advent of GPT-4 introduced more sophisticated methods like GPTScore \cite{fu2023gptscore}, utilizing its zero-shot capabilities for versatile text evaluation, and LLM-Eval \cite{llm-eval}, which employs a single LLM prompt to robustly evaluate conversational quality, correlating strongly with human judgments. Additionally, the LLM-as-a-judge approach \cite{llm-as-a-judge} finds GPT-4 approximating human evaluations with high agreement rates, offering a scalable alternative where traditional methods are impractical. These advancements highlight the increasing use of LLMs not only in generating text but also in evaluating it, a direction that aligns with this work and enhances the scalability, versatility, and precision of assessment methods.

\noindent\textbf{Human Preferences. }
The rule-based method was manually developed based on observations of the training set and incorporating human preferences via the $R_{basic}(S)$ reward model described in section~\ref{humanpref}. Our training set comprised 810 samples, with 296 of these identified as having paper structures. To detect these structures seen in 296 syntheses, we implemented a rule-based binary classifier using a set of predefined rules. This classifier utilized a paper structure vocabulary comprising 17 terms and 9 reference identifier regular expressions.
The vocabulary-based identifier achieved an F1-score of 67.80\%, while the reference identifier detection reached an F1-score of 90.87\%. When both identifiers were combined, the final model achieved an overall F1-score of 98.67\%, with only 10 misclassified cases. Further analysis involved annotating outputs from the \textit{SFT} and \textit{Vanilla} models and comparing them to our model's results. Human annotations indicated that the \textit{Vanilla} model had 31.62\% of its syntheses with paper structure, while the \textit{SFT} model had 2.99\%. Our model identified 32.47\% of the \textit{Vanilla} syntheses and 1.28\% of the \textit{SFT} syntheses as having paper structures. This demonstrates that our paper structure identifier model effectively contributes to the reward mechanism, with a focus on improving paper structure adherence in synthesis.


\noindent\textbf{Consistency Comparison. } 
In \autoref{consistency}, we analyzed GPT-4 evaluator behavior comparing the \textit{Vanilla} baseline model to the \textit{SFT+RLAIF (w/ GPT-4 Features)} optimized model output generated thrice on the same test set.
The results reveal that the \textit{Vanilla} model's output is inconsistent across three different runs.
This variability suggests that the \textit{Vanilla} model is less reliable as a synthesizer.
In contrast, the \textit{SFT+RLAIF} model exhibits much greater stability, outperforming the Vanilla model, with syntheses evaluations closely clustered and demonstrating consistent performance of syntheses evaluations. This enhanced consistency highlights the effectiveness of RLAIF in improving model reliability. 
The GPT-4 evaluator's ability to provide reliable and consistent assessments further supports these findings, making it an effective tool for evaluating model performance. Overall, the experiment underscores the benefits of RLAIF in delivering both consistent and higher-quality results.


\section{Conclusion}
\label{sec:conclude}
In conclusion, this work presents the LLMs4Synthesis framework, a comprehensive approach designed to enhance open-source LLMs for generating high-quality scientific syntheses. Addressing the challenges posed by the growing volume and complexity of scientific literature, the framework introduces new synthesis types, and quality evaluation criteria, and leverages RLAIF to improve synthesis generation. The study not only demonstrates the effectiveness of LLMs in producing and evaluating scientific summaries but also provides publicly available resources to further advance research in this domain, fostering greater accessibility and collaboration within the scientific community.

\begin{acks}
We thank \textit{Julia Evans} for her invaluable work on the "Human Evaluation of Synthesis Quality". This work is jointly supported by the \href{https://www.nfdi4datascience.de/}{NFDI4DataScience initiative} (DFG, German Research Foundation, Grant ID: 460234259) and the \href{https://scinext-project.github.io/}{SCINEXT project} (BMBF, German Federal Ministry of Education and Research, Grant ID: 01lS22070).
\end{acks}

\bibliographystyle{ACM-Reference-Format}
\bibliography{sample-base}




\end{document}